\documentclass[runningheads]{llncs}

\usepackage[T1]{fontenc}
\usepackage{graphicx}
\usepackage{color}
\usepackage{algorithm,algpseudocode}
\usepackage{amsmath}
\usepackage{amsfonts}
\usepackage{amssymb}
\usepackage{relsize}
\usepackage{hyperref}
\usepackage{multirow}
\usepackage{wrapfig}

\newcommand{\myx}{\mathbf{x}}
\newcommand{\relu}{{ReLU}}
\newcommand{\myinc}{\mathit{inc}}
\newcommand{\mydec}{\mathit{dec}}

\usepackage{tikz}
\usetikzlibrary{arrows,shapes,automata,petri,positioning,calc}

\tikzset{
    inc/.style={
        circle,
        thick,
        draw=black,
        fill=green!30,
        minimum size=5mm,
    },
    dec/.style={
        circle,
        thick,
        draw=black,
        fill=red!30,
        minimum size=5mm,
    },
    neu/.style={
        circle,
        thick,
        draw=black,
        fill=gray!30,
        minimum size=5mm,
    },
    state/.style={
        circle,
        thick,
        draw=blue!75,
        fill=blue!20,
        minimum size=5mm,
    },
}
\begin{document}

\title{Efficiently Finding Adversarial Examples with DNN Preprocessing}
%
%

\author{Avriti Chauhan\inst{1} \and
	Mohammad Afzal\inst{1,2} \and
	Hrishikesh Karmarkar\inst{1} \and
	Yizhak Yisrael Elboher\inst{3} \and
	Kumar Madhukar\inst{4} \and
	Guy Katz\inst{3}}
%
%
\institute{TCS Research, Pune, India \and
Indian Institute of Technology Bombay, Mumbai, India \and
Hebrew University of Jerusalem, Jerusalem, Israel \and
Indian Institute of Technology Delhi, Delhi, India
\email{\{avriti.chauhan,afzal.2,hrishikesh.karmarkar\}@tcs.com}\\
\email{\{yizhak.elboher,g.katz\}@mail.huji.ac.il}\\
\email{madhukar@cse.iitd.ac.in}}

\maketitle              
\begin{abstract}
	Deep Neural Networks (DNNs) are everywhere, frequently performing a
	fairly complex task that used to be unimaginable for machines to carry
	out. In doing so, they do a lot of decision making which, depending on
	the application, may be disastrous if gone wrong. This necessitates a
	formal argument that the underlying neural networks satisfy certain
	desirable properties. Robustness is one such key property for DNNs,
	particularly if they are being deployed in safety- or business-critical
	applications.  Informally speaking, a DNN is not robust if very small
	changes to its input may affect the output in a considerable way (e.g.
	changes the classification for that input). The task of finding an
	adversarial example is to demonstrate this lack of robustness, whenever
	applicable.  While this is doable with the help of constrained
	optimization techniques, scalability becomes a challenge due to
	large-sized networks. This paper proposes the use of information
	gathered by preprocessing the DNN to heavily simplify the optimization
	problem. Our experiments substantiate that this is effective, and does
	significantly better than the state-of-the-art.

\keywords{Adversarial Examples \and Constrained Optimization \and DNN Analysis}
\end{abstract}

\section{Introduction}
\label{sec:intro}

It would not be an inexcusable exaggeration, if one at all, to say that
Artificial Intelligence and Machine Learning touch every facet of our lives
today. While this makes us more capable, it also necessitates that we be
responsible in the use of these techniques and their artifacts, especially when
we deploy them in safety- or business-critical applications.  Consider a Deep
Neural Network (DNN) guiding a self-driving car, signalling it to stop, slow
down, or move at a traffic signal. We would like such a DNN to be trustworthy
and robust. For instance, the DNN must classify a stop signal correctly even on
a rainy day when the brightness is low. In fact, we would like to reason about
these formally, so that we can either produce an example which the DNN
misclassifies merely due to a small, insignificant change, or guarantee that
none exists. Such examples are called \emph{adversarial} examples, and they are
useful not just for improving the network (through adversarial training) but
also in deciding when the network should relinquish control to a more
dependable entity.

The problem of finding adversarial examples, to demonstrate lack of
robustness, has gained a lot of attention in the last several years. A number
of techniques have been developed for this task, both \emph{complete}
(e.g.~\cite{katz2017reluplex,huang2017safety}) and \emph{incomplete}
(e.g.~\cite{gehr2018ai2,singh2018boosting,wang2018formal}), trading off
scalability for precision and vice-versa. While we postpone the discussion of
strengths and limitations of these to the related work
(Sect.~\ref{sec:related}), the main challenge in this is to find the right
balance of efficiency and completeness, particularly for large networks.
This paper puts forth an approach that balances the two, by starting with an
extremely light-weight incomplete method which can be refined, layer-by-layer,
into a complete method.

In a recent work on abstraction-refinement of DNNs, Yizhak et
al.~\cite{elboher2020abstraction} proposed preprocessing of DNNs to gather
useful behavioral information of each neuron. In particular, they looked at how
an increase or decrease in the value of an intermediate-layer (or, internal)
neuron may affect (increase/decrease) the output values, and used this
information to merge \emph{similarly-behaving} neurons into one. The key
insight that we derive from their work is that this increment-decrement marking
is capable of telling us how we may bring about a misclassification (if one
exists) from \emph{any} layer. From the output layer, we know that we can get
it by \emph{decreasing} the current winning class, and \emph{increasing} the
runner-ups as much as possible. In the penultimate layer, because this layer is
also marked w.r.t. the output-marking, we know it is possible to bring about a
misclassification by increasing the increment neurons and decreasing the
decrement ones as much as possible. And, similarly, all the way to the input
layer -- we know that if we give the increment (decrement) input neurons as
high (low) a value as we can, then we will get an misclassification at the
output layer (and thus an adversarial example) if one exists.

This insight allows us to transform the problem of finding adversarial examples
into an optimization problem. But there are a few challenges that arise in
making this practicable. Firstly, many neurons may show a \emph{mixed}
behavior, and an increment/decrement marking may not be possible in such cases.
Secondly, between two increment neurons in the same layer, it is not clear
which one should be prioritized (for increasing) over the other. The same is
true for two decrement neurons. And even between an increment and a decrement
neuron, it is not clear whether increasing the former is more important than
lowering the latter, or the other way around. Essentially, a good objective
function for the optimization task is unknown. Lastly, if we are unable to find
an adversarial example at the first layer (or, for that matter, any layer
except the last), can we guarantee that an adversarial example does not exist.
This paper proposes the idea of using preprocessing information for finding
adversarial examples efficiently, and also discusses how the aforesaid
challenges may be addressed in this process.

In what follows, we cover the necessary background concept and present an
illustrative example in Sect.~\ref{sec:background}. Due to lack of space, a
formal presentation of our algorithm has been pushed to
Appendix~\ref{sec:method}. Sect.~\ref{sec:exps} shows the promise of our
approach with the results of our initial experiments on ACAS Xu benchmarks,
using a prototype implementation. We end with a discussion of the related and
potential future work.

\section{Background}
\label{sec:background}

A deep neural network (DNN) is described by an underlying weighted graph $D=(N,E,W)$, where $N$ is a
set of nodes and $E$ the set of edges. The set of nodes $N$ is partitioned into successive
\emph{layers}, $N_0, N_1, N_2, \ldots, N_k, N_{y}$, where layer $N_0$ is the
\emph{input layer}, $N_1, N_2, \ldots, N_k$ are \emph{hidden layers} and
$N_{y}$, $y=k+1$ is the \emph{output layer}. The nodes of the graph are
called \emph{neurons} and are ordered within a layer with the
$j^{th}$ neuron in the $i^{th}$ layer denoted as $n_{ij}$. 
Each neuron in $N_i$ is connected to neurons in $N_{i+1}$ by directed
edges from the edge set $E\subseteq N \times N$, which have a real valued weight $W:E \rightarrow \mathbb{R}$. 
When an input is applied to the neurons of $N_0$, every neuron $n_{ij}$, for $0 < i \leq k+1$ computes a weighted sum 
over values of neurons in the previous layer that are connected to it and adds a real valued \emph{bias} to it. 
Moreover, every neuron $n_{ij}$ in each hidden layer $N_i$, $1 \leq i \leq k$, applies an \emph{activation function} to the weighted sum. We assume that all activation functions are Rectified Linear Units ($\relu$), written as $ReLU(x)=max(0,x)$.
The DNN classifies an input vector $\myx$ applied to neurons of the input layer
$N_0$ based on a poll of the values of each output neuron $n_{yi} \in N_y$. Let
us call $n_{yi}$ as the winning neuron for input $\myx$ if it gets the highest
value among all output neurons. In this case, we call $i$ as the class of
$\myx$. 
Given a DNN $D$ and an input $\myx$ with winning neuron $n_{yi}$, an
adversarial input $\myx'$ is one that is within the allowed perturbation range
$\myx' = \myx \pm \delta$ such that its winning neuron is not $n_{yi}$.
\subsection{Illustrative Example}
\label{sec:example}
 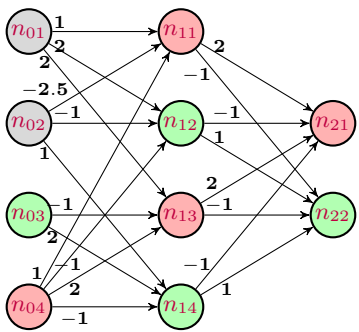
\begin{wrapfigure}{l}{0.50\textwidth} 
 \vspace{-\baselineskip}
 \begin{minipage}{0.5\textwidth}
	\centering
	\begin{tikzpicture}[node distance=0.6cm and 1.4cm,>=stealth',auto, every place/.style={draw},inner sep=1pt]
		\node [neu] (L1N1) {\textcolor{purple}{\footnotesize ${n_{01}}$}};
		\node [neu] (L1N2) [below=of L1N1] {\textcolor{purple}{\footnotesize $n_{02}$}};
		\node [inc] (L1N3) [below=of L1N2] {\textcolor{purple}{\footnotesize $n_{03}$}};
		\node [dec] (L1N4) [below=of L1N3] {\textcolor{purple}{\footnotesize $n_{04}$}};
		\node [dec] (L2N1) [right=of L1N1] {\textcolor{purple}{\footnotesize $n_{11}$}};
		\node [inc] (L2N2) [right=of L1N2] {\textcolor{purple}{\footnotesize $n_{12}$}};
		\node [dec] (L2N3) [right=of L1N3] {\textcolor{purple}{\footnotesize $n_{13}$}};
		\node [inc] (L2N4) [right=of L1N4] {\textcolor{purple}{\footnotesize $n_{14}$}};
		\node [dec] (L3N1) [right=of L2N2] {\textcolor{purple}{\footnotesize $n_{21}$}};
		\node [inc] (L3N2) [right=of L2N3] {\textcolor{purple}{\footnotesize $n_{22}$}};

		\path[->] (L1N1) edge node[above, xshift=-0.6cm, yshift=0.0cm] {{$\scriptstyle \mathbf{1}$}} (L2N1);
		\path[->] (L1N1) edge node[above, xshift=-0.6cm, yshift=0.3cm] {{$\scriptstyle\mathbf{2}$}} (L2N2);
		\path[->] (L1N1) edge node[above, xshift=-0.8cm, yshift=0.7cm] {{$\scriptstyle\mathbf{2}$}} (L2N3);

		\path[->] (L1N2) edge node[above, xshift=-0.8cm, yshift=-0.3cm] {$\scriptstyle \mathbf{-2.5}$} (L2N1);
		\path[->] (L1N2) edge node[above, xshift=-0.5cm, yshift=-0.0cm] {$\scriptstyle \mathbf{-1}$} (L2N2);
		\path[->] (L1N2) edge node[above, xshift=-0.8cm, yshift=0.7cm] {$\scriptstyle \mathbf{1}$} (L2N4);
		
		\path[->] (L1N3) edge node[above, xshift=-0.6cm, yshift=-0cm] {$\scriptstyle \mathbf{-1}$} (L2N3);
		\path[->] (L1N3) edge node[above, xshift=-0.7cm, yshift=0.2cm] {$\scriptstyle \mathbf{2}$} (L2N4);

		\path[->] (L1N4) edge node[above, xshift=-0.9cm, yshift=-1.5cm] {$\scriptstyle \mathbf{1}$} (L2N1);
		\path[->] (L1N4) edge node[above, xshift=-0.5cm, yshift=-0.8cm] {$\scriptstyle \mathbf{-1}$} (L2N2);
		\path[->] (L1N4) edge node[above, xshift=-0.4cm, yshift=-0.5cm] {$\scriptstyle \mathbf{2}$} (L2N3);
		\path[->] (L1N4) edge node[above, xshift=-0.4cm, yshift=-0.3cm] {$\scriptstyle \mathbf{-1}$} (L2N4);
%

		\path[->] (L2N1) edge node[above, xshift=-0.5cm, yshift=0.3cm] {$\scriptstyle \mathbf{2}$} (L3N1);
		\path[->] (L2N1) edge node[above, xshift=-0.8cm, yshift=0.5cm] {$\scriptstyle \mathbf{-1}$} (L3N2);

		\path[->] (L2N2) edge node[above, xshift=-0.4cm, yshift=0.0cm] {$\scriptstyle \mathbf{-1}$} (L3N1);
		\path[->] (L2N2) edge node[above, xshift=-0.5cm, yshift=0.3cm] {$\scriptstyle \mathbf{1}$} (L3N2);
		
		\path[->] (L2N3) edge node[above, xshift=-0.6cm, yshift=-0.3cm] {$\scriptstyle \mathbf{2}$} (L3N1);
		\path[->] (L2N3) edge node[above, xshift=-0.5cm, yshift=-0.0cm] {$\scriptstyle \mathbf{-1}$} (L3N2);
		
		\path[->] (L2N4) edge node[above, xshift=-0.8cm, yshift=-0.8cm] {$\scriptstyle \mathbf{-1}$} (L3N1);
		\path[->] (L2N4) edge node[above, xshift=-0.4cm, yshift=-0.5cm] {$\scriptstyle \mathbf{1}$} (L3N2);
	\end{tikzpicture}
    \end{minipage}
	\caption{Illustrative Example}\label{fig:simple}
    \end{wrapfigure}
Consider the DNN of Fig. \ref{fig:simple} with $\relu$ activation function and zero bias for each neuron. 
The perturbation for each input neuron is $\delta=0.5$.
On applying the input vector $\myx=\langle 0.6, -1.9, -0.7, -1\rangle$ 
the network produces the output $\langle 4.6, -0.25 \rangle$ with $n_{21}$ as the winning neuron. 
Any input $\myx'$ in the perturbation range $\myx \pm 0.5$ is adversarial if the winning neuron for it is $n_{22}$. 
For $n_{22}$ to become the winning neuron, its value needs to increase, while the value of $n_{21}$ needs to decrease. 
To keep track of neurons whose values need to either increase or decrease to effect a change in the winning neuron, we preprocess the DNN by adopting the neuron labelling scheme proposed by Elboher et. al.~\cite{elboher2020abstraction} with labels chosen from the set $\{\myinc, \mydec\}$.
The neuron $n_{22}$ is labelled $\myinc$ (colored \emph{green}), while $n_{21}$ is labelled $\mydec$ (colored \emph{red}). The labelling
scheme of~\cite{elboher2020abstraction}, is then applied one layer at a time from the output layer back to the input layer. The resultant labels
are shown in Figure~\ref{fig:simple} as node colors. However, in a departure from~\cite{elboher2020abstraction} our modified technique does not split neurons in the input layer if it can be labelled both $\myinc$ and $\mydec$, simultaneously. This is because, no neurons need to be tracked for their increment/decrement behaviour beyond the input layer. These neurons are colored \emph{gray} and are called \emph{mixed} neurons.
The inputs neurons $n_{03}$ and $n_{04}$ get labelled $\myinc$ and $\mydec$, suggesting that their values need to be increased and decreased, respectively from their current values given by $\myx$, to effect a change in the winning neuron. Since, any change in the value of these neurons must be within the limits $\myx \pm 0.5$, 
the value of $n_{03}$ is set to $-0.2=-0.7+0.5$ and the value of $n_{04}$ is set to $-1.5=-1-0.5$. 
But, values for the mixed neurons $n_{01}$ and $n_{02}$ cannot be similarly determined since they do not have a label and hence
the direction of value change is unclear. 
To find values for these neurons, we pose an optimization query $Q=maximize((n_{12} + n_{14})-(n_{11} + n_{13}))$ 
to an SMT solver along with an encoding $\pi_N$ of the DNN sub-graph structure for each neuron at layer $N_1$ and a set of value bound 
constraints $\Delta$
for the \emph{mixed} neurons $n_{01}$ and $n_{02}$.
For example, the sub-graph encoding at neurons $n_{11}$ and $n_{12}$ are 
$n_{11} = (n_{01}*1 + 0) + (n_{02}* (-2.5) + 0)$ and 
$n_{12} = (n_{01}*2 + 0) + (n_{02}* (-1) + 0)$ resp.
The value bound constraints $\Delta$ are encoded as 
$(0.6 - 0.5) \leq n_{01} \leq (0.6 + 0.5)$ and $(-1.9 - 0.5) \leq  n_{02} \leq (-1.9 + 0.5)$.
Thus, the choice of assignments for $n_{01}$ and $n_{02}$ are restricted to values that maximize 
the objective function of the optimization query, which in turn maximizes (minimizes) the values assigned to neurons labelled 
$\myinc$ ($\mydec$). 
The satisfying assignments for $n_{01}$ and $n_{02}$ give a new set of values $\myx'$ for the input neurons. 
We note that restricting the encoding to $N_0$ and $N_1$ limits the size of the query which is a key
strength of the technique and vastly improves the performance of the solver as we shall see later in Sect.\ref{sec:exps}.
For the example in Fig. \ref{fig:simple}, solving the query $\varphi=Q \wedge \pi_N \wedge \Delta$ 
with the Z3~\cite{z3solver} solver, results in 
an input assignment $\myx'=\langle 0.1, -1.4, -0.2, -1.5\rangle$ with the corresponding network output $\langle 1.1, 1\rangle$, which does not change the winning class. 
At this point, we iterate by first strengthening the query $\varphi$ with additional constraints, denoted $\alpha$, for each neuron in $N_1$. For the example these constraints are $n_{12} \geq 3.1$, $n_{14} \geq -0.3$, $n_{11}  \leq 2.1$, and $n_{13} \leq -2.6$, and require that $\myinc$ labelled neurons $n_{12}$ and $n_{14}$ 
and $\mydec$ labelled neuron $n_{11}$ and $n_{13}$ get values that are respectively $\geq$ and $\leq$
than the values they get because of the assignment $\myx'$.
However, these constraints are added as \emph{soft}
constraints for each neuron to prevent over-constriction of the search space. To preclude a potential consequence that
the solver returns the same input assignment $\myx'$ again, we add a constraint that \emph{blocks} $\myx'$.
With these additional constraints 
$Z3$ returns another input assignment $\myx''=\langle 0.85, -1.4, -0.2, -1.5\rangle$, with the corresponding output 
$\langle 1.1, 1.75\rangle$, which changes the winning neuron and implying that $\myx''$ is an adversarial example. 
A formal presentation of our algorithm has been pushed to
Appendix~\ref{sec:method} due to lack of space.

\section{Experiments}
\label{sec:exps}
We have implemented a prototype tool and compared our results, on ACAS Xu
(Airborne Collision Avoidance System) benchmarks~\cite{acasxubenchmarks}, with
that of $\alpha,\beta$-CROWN~\cite{alphabetacrown} and
Marabou~\cite{katz2019marabou}\footnote{We have submitted all the artifacts as
supporting documents along with the paper.}.  $\alpha,\beta$-CROWN is the
winner of the 2nd International Verification of Neural Networks Competition
(VNN-COMP 2021)~\cite{vnncomp2021report}, and Marabou is a popular SMT-based
tool. The ACAS Xu benchmarks contain 45 DNNs, each having 5 input neurons, 6
hidden layers with 50 neurons each, and 5 output neurons. We check the
robustness of these networks against 10 different properties as explained
in~\cite{katz2017reluplex} to find adversarial examples within allowed
perturbation ranges for the inputs.

\begin{table}
\caption{Comparison of our tool with $\alpha,\beta$-CROWN and Marabou}\label{resulttable}
\resizebox{\textwidth}{!}{\begin{tabular}{|cccccccc|}
\hline
\multicolumn{1}{|c|}{}                                      & \multicolumn{1}{c|}{}                                       & \multicolumn{2}{c|}{\textbf{\begin{tabular}[c]{@{}c@{}}$\alpha,\beta$-CROWN\\ (with Gurobi)\end{tabular}}} & \multicolumn{2}{c|}{\textbf{\begin{tabular}[c]{@{}c@{}}Marabou\\ (with Gurobi)\end{tabular}}} & \multicolumn{2}{c|}{\textbf{\begin{tabular}[c]{@{}c@{}}Our tool\\ (with z3 solver)\end{tabular}}} \\ \cline{3-8} 
\multicolumn{1}{|c|}{\multirow{-2}{*}{\textbf{Properties}}} & \multicolumn{1}{c|}{\multirow{-2}{*}{\textbf{\#instances}}} & \multicolumn{1}{c|}{\#violated}           & \multicolumn{1}{c|}{runtime(s)}                            & \multicolumn{1}{c|}{\#violated}      & \multicolumn{1}{c|}{runtime(s)}                        & \multicolumn{1}{c|}{\#violated}                          & runtime(s)                             \\ \hline
\multicolumn{1}{|c|}{Prop\_1}                               & \multicolumn{1}{c|}{45}                                     & \multicolumn{1}{c|}{0}                    & \multicolumn{1}{c|}{63.56}         & \multicolumn{1}{c|}{0}               & \multicolumn{1}{c|}{648.94}    & \multicolumn{1}{c|}{0}                                   & 83.78                                  \\ \hline
\multicolumn{1}{|c|}{Prop\_2}                               & \multicolumn{1}{c|}{45}                                     & \multicolumn{1}{c|}{38}                   & \multicolumn{1}{c|}{425.82}                                & \multicolumn{1}{c|}{35}              & \multicolumn{1}{c|}{1489.19}                           & \multicolumn{1}{c|}{34}                                  & 29.69                                  \\ \hline
\multicolumn{1}{|c|}{Prop\_3}                               & \multicolumn{1}{c|}{45}                                     & \multicolumn{1}{c|}{3}                    & \multicolumn{1}{c|}{93.26}                                 & \multicolumn{1}{c|}{3}               & \multicolumn{1}{c|}{306.31}                            & \multicolumn{1}{c|}{3}                                   & 83.42                                  \\ \hline
\multicolumn{1}{|c|}{Prop\_4}                               & \multicolumn{1}{c|}{45}                                     & \multicolumn{1}{c|}{3}                    & \multicolumn{1}{c|}{11.74}                                 & \multicolumn{1}{c|}{3}               & \multicolumn{1}{c|}{95.39}                             & \multicolumn{1}{c|}{3}                                   & 83.07                                  \\ \hline
\multicolumn{1}{|c|}{Prop\_5 to 10}                         & \multicolumn{1}{c|}{6}                                      & \multicolumn{1}{c|}{0}                    & \multicolumn{1}{c|}{578.19}                                & \multicolumn{1}{c|}{0}               & \multicolumn{1}{c|}{526.74}                            & \multicolumn{1}{c|}{0}                                   & 11.76                                  \\ \hline
\multicolumn{1}{|c|}{\textbf{Total}}                        & \multicolumn{1}{c|}{\textbf{186}}                           & \multicolumn{1}{c|}{\textbf{44}}          & \multicolumn{1}{c|}{\textbf{1172.57}}                      & \multicolumn{1}{c|}{\textbf{41}}     & \multicolumn{1}{c|}{\textbf{3066.57}}                  & \multicolumn{1}{c|}{\textbf{40}}                         & \textbf{291.72}                        \\ \hline
\multicolumn{8}{|l|}{\begin{tabular}[c]{@{}l@{}}\#instance: No. of benchmarks instances, \#violated: No. of violations/adversarial examples found, \\ runtime(s): Total tool execution time in seconds\end{tabular}}                                                                                                                                                                                                                   \\ \hline
\end{tabular}}
\end{table}

\vspace{-0.3in}

\subsubsection{Implementation}
\label{subsec:impl}

Our prototype tool is implemented in Python, and we have used Z3\cite{z3solver}
solver's python API, Z3Py (v4.8.15), for constraint solving. The tool takes 2
inputs, a feed-forward DNN (using $\relu$ activation function), and a standard
property file as used in VNN-COMP 2021 that describes the network's output
behavior like robustness w.r.t inputs or unchanged classification. Based of
this input property, we identify the increment/decrement output neurons, and
perform the neuron marking.
We have implemented the iterative algorithm as explained in section
\ref{sec:example} with the DNN sub-graph encoding restricted to the input layer
and the first layer. The number of iterations, chosen heuristically, was fixed
at 80. If the tool finds an adversarial example, it returns the input and
prints the corresponding network output. Otherwise, it returns ``unknown''.

\subsubsection{Results}
Table~\ref{resulttable} contains the results of our experiments. Properties 1
to 4 are applied on all 45 DNNS, whereas property 5 to 10 are applied on 6
separate ACAS Xu DNNs. We conducted these experiments on a machine with 16 GB
RAM, 3.60 GHz Intel processor, running Ubuntu 20.04, with a 116 seconds timeout
for each benchmark instance running on single core.

Our tool finds violations (adversarial inputs) in 40 among a total of 186 benchmark
instances. While comparing our tool with Marabou and $\alpha,\beta$-CROWN, we
have used their respective versions submitted to VNN-COMP
2021~\cite{vnncomp2021report}. The number of violations found by our tool is comparable to those found by Marabou, which
finds 41/186, and $\alpha,\beta$-CROWN which can find 44/186
violations. In terms of the total running time for all the 186 benchmarks, our
tool is $\approx$ 75\% faster compared to $\alpha,\beta$-CROWN, and $\approx$
90\% faster than Marabou.  A comparison of the respective running times for all
three tools on the 40 benchmarks for which we are able to find an adversarial
violation reveals that our tool finds them in just 9.2s and performs
significantly better in terms of total execution time compared to Marabou,
which takes 799.7s. We also manage to improve upon the 13.9s taken by
$\alpha,\beta$-CROWN for the same benchmark instances.
The improvement in running time comes from the fact that our approach
restricts the size of the problem/optimization instance to the first two
layers of the DNN.

\section{Related Work}
\label{sec:related}

In the past few years, there has been considerable work on proving adversarial robustness of deep neural networks. 
The vulnerability of deep neural networks to adversarial examples was first discovered in~\cite{szegedy2013intriguing}. Following this
there have been numerous subsequent results that study adversarial robustness using dynamic analysis techniques such as heuristic search ~\cite{carliniwagner,nguyen2015deep,tabacof2016exploring,goodfellow2014explaining,moosavisimple,gu2014towards}.
Another line of work poses adversarial robustness as a deep neural 
network verification problem and borrows tools and techniques from classical program verification for its solution. 
Among these, ~\cite{huang2017safety,bastani2016measuring,katz2017reluplex,ehlers2017formal} use constraint solving with various enhancements to enable an exhaustive search over the input perturbation interval for adversarial examples. Although, these techniques are complete they fail to scale as the network size increases. 
Abstraction based techniques work with an over-approximation of the network either by removing neurons~\cite{elboher2020abstraction} or 
by abstracting the state space computed by the network
~\cite{gehr2018ai2,singh2018fast,wang2018formal,tran2019star,yang2021improving,singh2018boosting,singh2019beyond,muller2022prima,wang2021beta,prabhakar2019abstraction}. Although, these methods scale better than constraint solving based exact techniques they are incomplete and 
suffer from the problem of false adversarial examples. Our work proposes to bridge the gap between these classes of techniques
by borrowing ideas from each viz. using constraint solving for optimization aided by behavioural abstraction of the DNN to efficiently
generate adversarial inputs.


\section{Conclusion and Future Work}
\label{sec:conc}

The ability to generate adversarial examples is crucial for robustness and
trustworthiness of DNNs, especially when they are used in safety-critical
application domains such as autonomous vehicles and precision medicine. This
paper presents an idea to find adversarial example efficiently. Our initial
experiments demonstrate that the approach has promise. As an immediate future
work, it would be worthwhile to complement the \emph{behavioral} marking of
neurons with other quantitative measures e.g. significance or importance of a
neuron, that can help obtain a good objective function for optimization.
Moreover, since the benefit of this approach comes from the reduction in
problem size, by shifting the problem to a layer closer to the input, it is
important to lay down necessary and sufficient conditions for moving the
optimization problem to a subsequent layer. It would also be interesting to
apply and tune this idea to work on a large class of benchmarks including
images, videos, and audio files.


%
%
%
\bibliographystyle{splncs04}
\bibliography{references}

\newpage

\appendix

\section{Algorithm}
\label{sec:method}
\begin{wrapfigure}{L}{0.50\textwidth}
\begin{minipage}{0.50\textwidth}
\vspace{-0.60in}
\begin{algorithm}[H]
\caption{Adversarial Generation}\label{alg:two}
\begin{algorithmic}[1]
\State {$\langle N_M, N_P\rangle \gets\mathit{IncDecAdv(D)}$} \label{incdecadv}
\For{each $n' \in N_P$}
\If{label of $n'$ is $\myinc$}
\State {set the value of $n'$ to $\myx(n')+\delta$}
\ElsIf{label of $n'$ is $\mydec$}
\State {set the value of $n'$ to $\myx(n')-\delta$}
\EndIf
\EndFor
\State {$\varphi \gets \mathit{Encode}(D,Q,\pi_N, \Delta)$}\label{encode}
\State{$it \gets 1$}
\While{$it \leq C$}
\State{$\myx'=\mathit{Solve}(\varphi)$} \label{solve}
\If{$x'$ is adversarial}
\State \Return {$x'$}
\EndIf
\State{$\varphi\gets \mathit{AddSoftConstraint}(\varphi,\alpha)$}\label{addlabelconstraint}
\State{$\varphi\gets \mathit{BlockExample}(\varphi,\myx')$}\label{blockexample}
\State{$\myx=\myx'$}
\State{$it \gets it+1$}
\EndWhile
\State \Return \emph{unknown}
\end{algorithmic}
\end{algorithm}
\end{minipage}
\vspace{-0.30in}
\end{wrapfigure}
For a DNN $D$, let $IncDecAdv(D)$ be the modified labelling scheme described in Sect. \ref{sec:example}. 
A call to $IncDecAdv(D)$ (Line \ref{incdecadv}) returns the set of \emph{mixed} input neurons $N_E$ and the set of labelled input neurons
$N_P$. If an input neuron $n' \in N_P$ has label $\myinc$ ($\mydec$), then its value is set to $\myx(n')+\delta$ ($\myx(n')-\delta$)
where $\myx(n')$ is the input value to neuron $n'$. A call to $\mathit{Encode}(D,Q,\pi_N, \Delta)$ (Line \ref{encode}) generates the query 
$\varphi$, where $Q$ is the optimization query, $\pi_N$ is the sub-graph structure encoding for nodes in $N_1$ and $\Delta$ are the value bound
constraints for the $\emph{mixed}$ neurons. The procedure then iterates for a pre-defined iteration count $C$. The query $\varphi$ is solved by 
the solver to give a new input assignment $\myx'$ (Line \ref{solve}). If $\myx'$ is adversarial, then the search exits successfully, else
$\varphi$ is strengthened with soft constraints $\alpha$ and the blocking constraint on $\myx'$ via calls to 
$\mathit{AddSoftConstraint}(\varphi,\alpha)$ (Line \ref{addlabelconstraint}) and $BlockExample(\varphi,\myx')$ (Line \ref{blockexample}) before
the solver is called again.
We note that the maximization function that is used in the optimization query $Q$ is only 
but one choice for an objective function. Algorithm \ref{alg:two} can work seamlessly with any other objective function that 
can better constrain $\varphi$ to improve the search for an adversarial example. 
The iterations continue till an adversarial input is generated or a fixed number of iterations is exhausted within a preset time-out duration for running the algorithm. 
\subsection{A note on completeness of the algorithm}
We would like to record that the technique described by Algorithm \ref{alg:two} can be modified into a complete procedure for generating adversarial inputs by - encoding the sub-graph of the DNN between the input layer $N_0$ and the output layer $N_y$, which in effect is the entire graph of the DNN and generating the query $Q$ over neurons in $N_y$. Even the \emph{soft} constraints can be written over neurons from $N_y$. This query $\varphi_{0y}$ when solved gives an input assignment that changes the winning neuron and hence is adversarial. 
The downside of this approach is that the algorithm does not scale and in essence is no better than existing techniques that take the full DNN into consideration. However, a more balanced approach generates queries $\varphi_{0i}$, for $0 < i \leq y$ at each layer starting from 
$N_1$ to $N_{y}$ and solves it for a fixed number of iterations. This also gives us a way to 
tune the generalised version of the algorithm for a trade-off between scalability, which solves queries $\varphi_{0i}$ over layers closer to the input layer and completeness, which solves the query $\varphi_{0y}$ at the output layer.

\end{document}